\documentclass[letterpaper]{article} 
\usepackage{aaai25}  
\usepackage{times}  
\usepackage[table]{xcolor} 
\usepackage{helvet}  
\usepackage{courier}  
\usepackage[hyphens]{url}  
\usepackage{graphicx} 
\usepackage{natbib}  
\usepackage{caption} 
\usepackage[most]{tcolorbox}
\usepackage{array}
\usepackage{algorithm}
\usepackage{algorithmic}

\usepackage{newfloat}
\usepackage{multirow} 
\nocopyright
\usepackage{listings}
\DeclareCaptionStyle{ruled}{labelfont=normalfont,labelsep=colon,strut=off} 
\floatstyle{ruled}
\newfloat{listing}{tb}{lst}{}
\floatname{listing}{Listing}
\title{ProRAC: A Neuro-symbolic Method for Reasoning about Actions with LLM-based Progression}
\author{
    Haoyong Wu, Yongmei Liu\thanks{Corresponding author.} 
}
\affiliations{
    Dept. of Computer Science, Sun Yat-sen University\\

    Guangzhou 510006, China\\
    wuhy259@mail2.sysu.edu.cn; ymliu@mail.sysu.edu.cn
}

\usepackage{bibentry}

\begin{document}

\maketitle

\begin{abstract}
Reasoning about actions and change (RAC) plays an important role in AI. It involves reasoning about preconditions and effects of actions, and has applications in planning, an important research area in classic AI. 
 In recent years, large language models (LLMs) have made remarkable progress in NLP and related fields, and reasoning over natural language has received much attention. Several benchmarks for RAC over natural language such as TRAC, ActionReasoningBench and ACPBench have been proposed. Evaluation of LLMs on these benchmarks demonstrates that LLMs face significant challenges in RAC. However, methods for improving RAC abilities of LLMs remain largely unexplored.
In this paper, we propose \textbf{ProRAC} (\textbf{P}rogression-based \textbf{R}easoning about \textbf{A}ctions and \textbf{C}hange), a neuro‐symbolic framework that leverages LLMs to tackle RAC problems. ProRAC extracts fundamental RAC elements including actions and questions from the problem, progressively executes each action to derive the final state, and then evaluates the query against the progressed state to arrive at an answer. We evaluate ProRAC on several RAC benchmarks, and the results demonstrate that our approach achieves strong performance across different benchmarks, domains, LLM backbones, and types of RAC tasks.

\end{abstract}

%

\section{Introduction}
Reasoning about actions and change (RAC) plays an important role in artificial intelligence. It involves reasoning about the preconditions and effects of actions. Intelligent agents must act on the world: 
taking purposeful actions, predicting the expected effects of such actions, and composing actions together to achieve complex goals. The last task is planning, it is an application of RAC, and an important research area in classical AI. 
There are three challenging problems in RAC: the Frame Problem  \cite{FRAME-PROBLEM}, the Ramification Problem \cite{RAMIFICATION}, and the Qualification Problem  \cite{QUALIFICATION}. The Frame Problem refers to determining what remains unchanged after an action is executed. The Ramification Problem concerns the many indirect effects that an action may have, which are not explicitly specified but result from underlying domain constraints. The Qualification Problem arises from the fact that the execution of an action usually depends on many preconditions, most of which are difficult to enumerate exhaustively.

In classical AI, many formalisms have been proposed for RAC, including Situation Calculus \citep{SITUATION-CALCULUS}, Event Calculus \citep{EVENT-CALCULUS}, and Action Language $\mathcal{A}$ \citep{ACTION-LANGUAGE-A}. Two fundamental reasoning methods to address RAC problems within these formalisms are Regression and Progression  \cite{SITUATION-CALCULUS}. Roughly, regression reduces a query about the future to a query about the initial state. Progression, on the other hand, changes the initial state according to the effects of each action and then checks whether the formula holds in the resulting state. One advantage of progression compared to regression is that after a state has been progressed, many queries about the resulting state can be processed without extra overhead. Moreover, when the action sequence becomes very long, regression becomes unmanageable.

\begin{figure*}[t]
    \centering
    \includegraphics[scale=0.90]{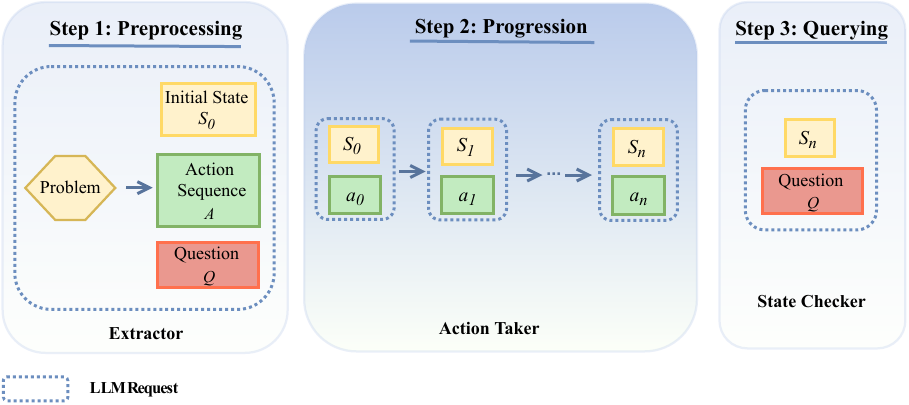}
    \caption{Overview of ProRAC, consisting of three steps: Preprocess, Progression and Querying.}
    \label{fig:framework}
\end{figure*}

While RAC problems have been extensively studied within classical symbolic approaches, large language models have demonstrated remarkable capabilities in recent years, offering new possibilities for solving RAC problems in natural language. Recently, a number of benchmarks for RAC over natural language have been proposed. \citet{TRAC} introduced TRAC, a benchmark targeting RAC problems that includes some basic tasks such as projection, executability, plan verification and goal recognition. However, TRAC  focuses solely on the classical planning domain Blocksworld.
Another RAC benchmark is ActionReasoningBench \citep{ACTIONREASONINGBENCH}, which covers multiple classical planning domains, such as Depots and Grippers. 
ACPBench \cite{ACPBENCH} and ACPBench Hard \cite{ACPBENCH-HARD} are two benchmarks in parallel with ActionReasoningBench. ACPBench also covers multiple planning domains and features both multiple-choice and true/false questions. ACPBench Hard is a generative version of ACPBench with open-ended questions. 

The success of LLMs in various reasoning tasks has also led to growing interest in neuro-symbolic AI approaches. However, neuro-symbolic AI goes beyond a mere combination of neural networks and symbolic solvers. In fact, neuro-symbolic approaches encompass a spectrum of integration paradigms, ranging from neural networks processing symbolic representations to systems where symbolic knowledge guides neural learning, and even purely neural systems that exhibit symbolic-like reasoning behaviors \cite{kautz}. This broader perspective recognizes that the boundary between neural and symbolic processing is more nuanced than traditional categorizations suggest.

While several benchmarks have been introduced, comprehensive neuro-symbolic frameworks for solving RAC problems remain scarce. LLM+AL \citep{LLM+AL} is a neuro-symbolic framework leveraging LLMs and symbolic solvers to solve RAC problems. The LLM functions as a semantic parser, translating problems expressed in natural language into an action language. Then a symbolic solver performs reasoning to produce the final solution based on translation results. 
While effective on planning tasks, the framework  requires human intervention, as in some cases the LLM consistently fails to accurately translate problems into a formal language. Besides, it only evaluates a few classical planning problems and does not involve the aforementioned benchmarks.

Beyond symbolic translation approaches, recent work has explored leveraging LLMs' inherent reasoning capabilities for progression-based planning and state transitions. Tree of Thoughts \citep{TOT} enables LLMs to deliberate over coherent units of text that serve as intermediate steps toward problem solving, essentially progressing through a search space of reasoning states. \citet{RAP} treats LLMs as world models that can simulate state transitions and uses Monte Carlo Tree Search to explore reasoning trajectories. These methods demonstrate that LLMs can effectively perform progression-based reasoning without explicit symbolic translation. However, these approaches have primarily targeted general reasoning tasks rather than systematic evaluation on RAC problems with established benchmarks, highlighting the need for systematic frameworks that leverage LLM-based progression for RAC problems.

To enhance the reasoning capabilities of LLMs in RAC problems, we propose \textbf{ProRAC}, a neuro-symbolic framework consisting of three steps — \textit{Preprocessing}, \textit{Progression}, and \textit{Querying}, all performed using LLMs. First, we extract initial state, action sequence, and the query from the problem described in natural language. 
Then, we iteratively call an LLM to execute each action in the action sequence based on the current state, updating the state accordingly. 
Finally, we use the LLM to evaluate the query against the progressed final state to generate answers.

In order to validate the effectiveness of ProRAC, we conducted experiments on three RAC benchmarks: TRAC, ACPBench, and ActionReasoningBench. We selected 14 different domains including Blocksworld, Depots, and Grippers from these benchmarks, and tested our approach using GPT-4o, GPT-4o-mini, and DeepSeek-v3 as underlying models. Experimental results demonstrate that our method outperforms directly feeding the problem description to LLMs. To the best of our knowledge, ProRAC is the first framework that can efficiently solve problems in the three RAC-oriented benchmarks. 

Furthermore, our analysis reveals that the errors made by LLMs on RAC tasks are closely related to the classic AI challenges of Frame, Ramification, and Qualification Problem, indicating that these long-standing challenges remain critical obstacles in current LLM reasoning capabilities.

To summarize, our main contributions are:
\begin{itemize}
    \item We propose ProRAC, a neuro-symbolic framework leveraging LLMs targeting RAC problems without relying on symbolic solvers.
    \item We conducted experiments on three RAC-oriented benchmarks across domains and tasks, demonstrating strong performance.
    \item Our study underscores that RAC, together with the associated classic AI challenges of Frame, Ramification, and Qualification Problem, continues to be of critical theoretical and practical importance in the era of LLMs.

\end{itemize}

\section{Related Work}

\begin{figure*}[t]
    \centering
    \includegraphics[scale=0.91]{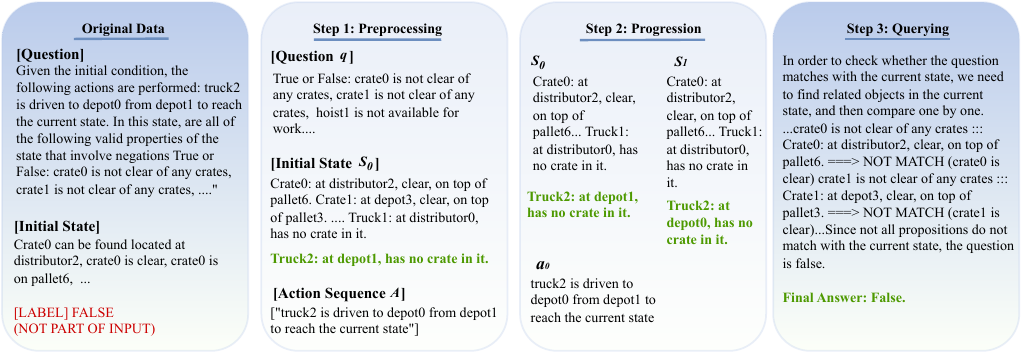}
    \caption{An illustrative example of ProRAC. The original data is processed step by step from Step One to Step Three.}
    \label{fig:example}
\end{figure*}

In classic AI, researchers addressed RAC problems by translating them into formal representations and employing symbolic solvers for reasoning. Various logical formalisms have been proposed for RAC, including Situation Calculus \citep{SITUATION-CALCULUS} and Event Calculus \citep{EVENT-CALCULUS}, as well as Action Languages $\mathcal{A,B,C}$ \citep{ACTION-LANGUAGE-A,ACTION-LANGUAGE-B, ACTION-LANGUAGE-C}, among others.

Although symbolic solver-based approaches are capable of solving RAC problems, such methods inevitably suffer from two major issues. First, symbolic methods often require manually translating problems into formal languages, this formalization process is expensive and demands a high level of expert knowledge from translators. Second, in symbolic methods, actions are strictly defined, which limits the generalization ability of symbolic solvers and makes them hard for handling 
varied natural language expressions. 


As LLMs have exhibited powerful semantic parsing capabilities, many researchers have proposed combining LLMs with symbolic solvers. 
In this neuro-symbolic paradigm, LLMs serve as translation modules that translate natural language described problems into formal representations, which are subsequently solved by symbolic solvers. 
This approach, known as autoformalization, has inspired many studies \citep{LOGIC-LM, LINC, VERUS-LM}. However, the ability of LLMs to accurately translate problems from natural language into formal languages remains limited.  Many works based on autoformalization demonstrate that LLMs still face significant challenges in the formalization process, particularly on complex reasoning problems \citep{AR-LSAT}.

Besides, it is worth mentioning that some studies \citep{QuaSAR,SYMBOLIC-COT} use LLMs as both translator and solver, where translation results are directly passed to LLMs to perform reasoning. This approach does not involve symbolic solvers, thus avoiding the translation challenges inherent in autoformalization while achieving greater robustness and generalization. We consider these works also to fall under the category of neuro-symbolic methods, as they follow the "formalize-solve-query" pipeline of neuro-symbolic approaches.

In addition to autoformalization, search-based methods represent another crucial direction for enhancing the reasoning capabilities of LLMs. These approaches leverage classical search algorithms such as A* search \cite{LLM-A*} and Monte Carlo Tree Search \cite{LLM+MCTS} to systematically explore the solution space, thereby improving the accuracy and robustness of LLM reasoning. In addition, recent studies have proposed combining search paradigms with prompting strategies \cite{TOT, GOT}, integrating structured exploration with prompt engineering to further strengthen the reasoning performance of LLMs.

\section{Overview of ProRAC}
\begin{table}[t]
\centering
\setlength{\tabcolsep}{0.95pt} 
\renewcommand{\arraystretch}{1.25} 

\begin{tabular}{|>{\centering\arraybackslash}m{0.34\linewidth}|>{\centering\arraybackslash}m{0.21\linewidth}|>{\centering\arraybackslash}m{0.21\linewidth}|>{\centering\arraybackslash}m{0.21\linewidth}|}
\hline
Model & PR & EXE & PV\\ \hline
4o & 94.73 & 96.58 & 88.69 \\
4o-0CoT & 96.19 & 97.22 & 87.51 \\
4o-2CoT & 97.77 & 95.56 & 93.33 \\
4o-SC&97.78&95.56&93.33\\
\textbf{ 4o-ProRAC} & \textbf{100} & \textbf{100} & \textbf{100} \\ \hline
4o-mini & 95.56 & 73.33 & 88.89 \\
4o-mini-0CoT & 95.56 & 68.88 & 84.44 \\
4o-mini-2CoT & 93.33 & 73.33 & 86.67 \\
4o-mini-SC&88.89&84.44&91.11\\
\textbf{4o-mini-ProRAC }& \textbf{100} & \textbf{100} & \textbf{100} \\ \hline
v3 & 98 & 97.77 & 95.56 \\
v3-0CoT & 100 & 97.77 & 100 \\
v3-2CoT & 100 & 100 & 95.56 \\
v3-SC&97.78&97.78&100\\
\textbf{ v3-ProRAC }& \textbf{100} & \textbf{100} & \textbf{100} \\ \hline
\end{tabular}
\caption{Performance (\%) of ProRAC and baseline methods on TRAC benchmark. PR = Projection, EXE = Executability, PV = Plan Verification.}
\label{tab:full-results-TRAC}
\end{table}

\begin{table*}[t]
\centering
\renewcommand{\arraystretch}{1.2}  
\footnotesize  
\setlength{\tabcolsep}{2.5pt}  
\resizebox{\textwidth}{!}{
\begin{tabular}{|c|cccc|cccc|cccc|cccc|}
\hline
\multirow{2}{*}{\scriptsize Model} & \multicolumn{4}{c|}{\scriptsize Depots} & \multicolumn{4}{c|}{\scriptsize Driverlog} & \multicolumn{4}{c|}{\scriptsize Mystery} & \multicolumn{4}{c|}{\scriptsize Grippers} \\
\cline{2-17}
 & \scriptsize AE & \scriptsize EFF & \scriptsize FT & \scriptsize ST & \scriptsize AE & \scriptsize EFF & \scriptsize FT & \scriptsize ST & \scriptsize AE & \scriptsize EFF & \scriptsize FT & \scriptsize ST & \scriptsize AE & \scriptsize EFF & \scriptsize FT & \scriptsize ST \\
\hline
4o & \textbf{97.74} & 79.1 & 78.13 & 73.68 & 77.14 & 80.95 & 70 & 80 & 75 & 45.5 & 89.48 & 74.1 & 95.24 & 90.38 & 95.92 & 83.33 \\
4o-0CoT & 97.37 & 48.84 & 84.38 & 39.47 & 82.86 & 90.95 & 76.67 & 80 & 82.1 & 54.55 & 88.9 & 64 & 95.24 & 90.38 & 93.88 & 83.33 \\
4o-2CoT & 97.37 & 74.41 & 75 & 78.94 & 94.28&90.47&73.33&\textbf{92} & 80 & 52.2 & 79 & 69 & 95.24 & 89 & 90 & 83.3 \\
4o-SC&97.37&69.77&75&81.58&91.42&90.47&90&92&75&43.47&84.21&65.51&95.24&88.46&93.87&90\\
\textbf{ 4o-ProRAC} & 92.11 & \textbf{90.7} & \textbf{93.75} & \textbf{94.74} & \textbf{97.14} & \textbf{95.24} & \textbf{93.33} & 82.6 & \textbf{90} & \textbf{91.3} & \textbf{100} & \textbf{86.24} & \textbf{95.24} & \textbf{97.73} & \textbf{97.62} & \textbf{90} \\
\hline
4o-mini & \textbf{94.7} & 58.14 & 67.74 & 81.6 & 77.14 & 90.5 & 80 & 56 & 65 & 52.2 & 68.42 & 58.62 & 92.9 & 84.62 & 89.6 & 66.67 \\
4o-mini-0CoT & 89.5 & 53.5 & \textbf{71} & 79 & 71.43 & \textbf{95.24} & 86.67 & \textbf{84} & 65 & 39.13 & 84.21 & 55.17 & 83.33 & \textbf{86.54} & 85.71 & 70 \\
4o-mini-2CoT & 92.1 & \textbf{76.64} & 65.62 & 60.52 & 85.71&85.72&\textbf{90}&64& 70 & 39.13 & 79 & \textbf{59} & \textbf{91} & 79 & 88 & 70 \\
4o-mini-SC&89.47&51.16&71.87&\textbf{86.84}&\textbf{80}&95.23&90&84&72.5&39.13&73.68&51.72&92.85&84.61&\textbf{89.79}&\textbf{73.33}\\
\textbf{4o-mini-ProRAC }& 86.84 & 44.19 & 59.4 & 65.79 & 74.29 & 85.71 & 86.67 & 68& \textbf{75} & \textbf{65.22} & \textbf{79} & 58.62 & 88.1 & 59.62 & 87.76 & 70 \\
\hline
v3 & \textbf{97.4} & 81.4 & 65.6 & 76.32 & 85.71 & 80.95 & 76.67 & 76 & 77.5 & 52.2 & 84.21 & 75.9 & 95.24 & 90.38 & 98 & 80 \\
v3-0CoT & 94.74 & 76.74 & 75 & 65.79 & 80 & 80.95 & 73.33 & 80 & 77.5 & 52.2 & 94.74 & 79.3 & 95.24 & 88.46 & 97.96 & 86.97 \\
v3-2CoT & 97.36 & 83.72 & 75 & 57.89 & 97.14&100&73.33&64& 80 & 91.3 & 94.74 & 76 & 95.24 & 88.46 & 100 & 83.33 \\
v3-SC&92.10&79.06&75&\textbf{86.48}&97.14&90.47&83.33&\textbf{96}&80&52.17&89.47&68.96&95.23&88.46&93.61&72.22\\
\textbf{ v3-ProRAC} & 92.11 & \textbf{86.05} & \textbf{87.5} & 81.6 & \textbf{100} & \textbf{100} & \textbf{90} & 88 & \textbf{95} & \textbf{100} & \textbf{100} & \textbf{100} & \textbf{97.61} & \textbf{100} & \textbf{100} & \textbf{100} \\
\hline
\end{tabular}
}
\caption{Performances(\%) of ActionReasoningBench on Depots, Driverlog, Mystery and Grippers. AE=Action Executability, EFF=Effects, FT=Fluent tracking, ST=State Tracking.}
\label{tab:full-result-ARB-1}
\end{table*}
An overview of the ProRAC framework is shown in Figure \ref{fig:framework}, consisting of three steps: Preprocessing, Progression and Querying. All three steps are carried out by  large language models, without relying on any symbolic solvers. In each step, manually crafted examples are introduced to make sure we can get the intended answer from LLMs. All examples are synthesized to make sure there is no data leakage in the prompt.

\subsection{Preprocessing}
The first step of ProRAC is preprocessing. In this step we prompt the LLM with the problem to extract basic RAC elements including: initial state $S_0$, action sequence $A$, and the query $q$, which is used in the final step to generate the answer. To ensure structured outputs from the LLM, we also include two manually crafted examples in the prompt as demonstrations.

\subsection{Progression}
Once step one is completed, we proceed to the next step, which is the core of ProRAC. In this step, the current states $S_i$ and each action $a_i$ in action sequence $A$ are sequentially provided to the LLM to generate the corresponding new states $S_{i+1}$, as RAC problems require temporal reasoning over actions and states. This process is performed iteratively and progressively, and we explicitly instruct the LLM to return all objects' states in the response. During this process, each action is executed by making a separate call to the LLM. For example, if $A$ contains $k$ actions: $a_0, a_1,..., a_{k-1}$, then executing every action in $A$ will produce $S_0, S_1,...,S_k$, where each $S_i$ represents the states of all objects in the problem. 

In addition to the current state and an action, we also include a manually crafted domain description and several examples in the prompt. 
Similar to definitions in PDDL or STRIPS, the domain description here specifies preconditions and effects of each action within the domain, and serves as domain knowledge accessible to the LLM. However, unlike the formal representations in PDDL or STRIPS, the domain description here is expressed in natural language, which enhances readability and reduces difficulty for users writing domain knowledge. 

Besides, it is possible that certain actions in the given sequence cannot be executed because their preconditions are not satisfied in the current state.
We add an additional executability check before taking each action. The executability check asks the LLM whether all preconditions of the action are satisfied in the current state. 

\subsection{Querying}
After step two, we obtain the final state $S_n$. In the third step, we ask the LLM whether $S_n$ satisfies the query $q$ obtained from step one. The query may contain multiple propositions, and the LLM return true only if $S_n$ satisfies all propositions in $q$. If there is any proposition cannot be satisfied, the LLM returns false as the answer. That is, the LLM returns True only if $q \subseteq S_n$.

\section{Experiments and Results}

\subsection{Benchmarks}
We evaluated ProRAC on three RAC-oriented benchmarks. We selected diverse domains from different benchmarks, ensuring that these domains didn't overlap with each other. These domains cover a wide range of scenarios, including logistics and transportation, block stacking, and grid-based navigation. Besides, since ProRAC is designed for RAC tasks rather than plan generation, which usually requires systematic search, we focused on tasks from these benchmarks which don't involve planning generation.

\begin{table*}[t]
\centering
\renewcommand{\arraystretch}{1.2}  
\footnotesize  
\setlength{\tabcolsep}{2.5pt}  
\begin{tabular}{|c|cccc|cccc|cccc|}
\hline
\multirow{2}{*}{\scriptsize Model} & \multicolumn{4}{c|}{\scriptsize Satellite} & \multicolumn{4}{c|}{\scriptsize Spanner} & \multicolumn{4}{c|}{\scriptsize Visitall} \\
\cline{2-13}
 & \scriptsize AE & \scriptsize EFF & \scriptsize FT & \scriptsize ST & \scriptsize AE & \scriptsize EFF & \scriptsize FT & \scriptsize ST & \scriptsize AE & \scriptsize EFF & \scriptsize FT & \scriptsize ST \\
\hline
4o & 75.68 & 79.12 & 78.72 & 66.77 & 75 & 84.48 & 85 & 72.22 & 69.23 & 86.05 & \textbf{93.33} & 80 \\
4o-0CoT & 81.08 & 83.33 & 85.11 & 66.67 & 69.44 & 74.07 & 83 & 75 & 64.1 & 83.72 & 93.33 & \textbf{86.67} \\
4o-2CoT & 75.67 & 83.3 & 87.23 & 75 & 94.44&88.88&93.61&72.22& 64.1 & 90.69 & 93.33 & 86.67 \\
4o-SC&78.37&83.33&91.48&75&80.55&81.48&91.48&75&74.36&83.72&93.33&93.33\\
\textbf{4o-ProRAC} & \textbf{94.6} & \textbf{100} & \textbf{93.62} & \textbf{83.8} & \textbf{88.89} & \textbf{92.6} & \textbf{100} & \textbf{94.44} & \textbf{94.87} & \textbf{95.35} & 86.67 & 53.33 \\
\hline
4o-mini & 56.8 & 62.5 & 72.34 & 61.11 & 77.8 & 74.1 & 89.4 & \textbf{82.86} & 66.67 & \textbf{74.42} & 86.67 & 66.67 \\
4o-mini-0CoT & 51.35 & 66.67 & 78.23 & 68.6 & 77.14 & 84.62 & 89.4 & 71.43 & 66.67 & 74.42 & 90 & \textbf{73.33} \\
4o-mini-2CoT & 75.67 & 62.5 & 78.72 & 66.67 & 85.71&\textbf{85.71}&\textbf{90}&64& 53.84 & 69.76 & \textbf{90} & 60 \\
4o-mini-SC&70.27&70.83&74.46&69.44&88.88&81.48&80&69.44&51.28&72.09&80&73.33\\
\textbf{4o-mini-ProRAC} & \textbf{92} & \textbf{83.33} & \textbf{83} & \textbf{77.78} & \textbf{93.33} & 66.37 & 75.56 & 80.56 & \textbf{79.49} & 65.12 & 80 & 60 \\
\hline
v3 & 86.49 & 70.83 & 85.11 & 63.9 & 83.33 & 88.46 & 92.48 & \textbf{78.79} & 76.92 & 90.7 & 93.33 & 73.33 \\
v3-0CoT & 83.78 & 70.83 & 80.85 & 61.11 & 82.86 & 88.46 & 91.3 & 64.71 & 76.92 & 93.02 & 86.67 & \textbf{86.67} \\
v3-2CoT & 83.78 & 75 & 83 & 66.67 & 100&88.88&91.48&66.66& \textbf{89.74} & 72.1 & 93.33 & 53.33 \\
v3-SC&83.78&75&82.97&75&80.55&88.88&93.61&72.22&74.35&90.69&93.33&86.67 \\
\textbf{v3-ProRAC }& \textbf{97.3} & \textbf{100} & \textbf{92.9} & \textbf{78.26} & \textbf{97.22} & \textbf{96.29} & \textbf{97.9} & 70.6 & 84.62 & \textbf{97.7} & \textbf{100} & 69.23 \\
\hline
\end{tabular}
\caption{Performances(\%) of ActionReasoningBench on Satellites, Spanner and Visitall. AE=Action Executability, EFF=Effects, FT=Fluent tracking, ST=State Tracking.}
\label{tab:full-result-ARB-2}
\end{table*}

\textbf{TRAC}. This benchmark covers the Blocksworld domain. We select projection, executability and plan verification tasks for evaluation. The projection task requires LLMs to evaluate whether all propositions in the query match the final state after taking the entire action sequence. The executability requires LLMs to assess whether all preconditions of an action are satisfied in the current state. For the plan verification, LLMs are prompted to check whether the given action sequence can reach the goal in the query.

\textbf{ActionReasoningBench}. We evaluated all domains in the benchmark including Depots, Driverlog, Mystery, Grippers, Satellite, Spanner, and Visitall. We selected action executability, effects, fluent tracking and state tracking tasks for evaluation. In this benchmark, the action executability task is the same as the executability task in TRAC. Besides, effects, fluent tracking and state tracking are also equivalent to the projection task in TRAC. However, the number of propositions in the benchmark's queries is much larger than those in TRAC.

\textbf{ACPBench}. Considering some domains in the benchmark overlap with those in ActionReasoningBench, we selected six domains, including Ferry, Swap, Logistics, Rovers, Floortile and Goldminer. We selected applicability, progression and validation tasks for evaluation. The applicability and progression tasks in the benchmark are equivalent to the executability and projection tasks in TRAC, respectively. While the validation task is conceptually similar to TRAC’s plan verification task, it offers a more fine-grained evaluation by categorizing action sequences into three distinct types: valid, applicable, or constituting a plan. As a result, the validation task presents a higher level of difficulty compared to the plan verification task in TRAC.

\subsection{Models and methods}
We used GPT-4o \cite{GPT-4o}, GPT-4o-mini, and DeepSeek-V3 \cite{V3} as the underlying models for evaluation. For comparison, we adopted Zero-Shot, Zero-Shot-CoT \cite{0shotCOT}, and Two-Shot-CoT \cite{FewshotCOT} prompting as baselines. In Two-Shot-CoT prompts, we followed a process similar to ProRAC: first identifying the corresponding objects, then checking the executability of actions, subsequently executing the actions to obtain the final state, and finally comparing the query with the final state to determine the answer.

\begin{table*}
\centering
\renewcommand{\arraystretch}{1.3}  
\resizebox{1.0\linewidth}{!}{%
\LARGE
\begin{tabular}{|>{\centering\arraybackslash}m{0.29\linewidth}|ccc|ccc|ccc|ccc|ccc|ccc|} 
\hline
\multirow{2}{*}{\centering\LARGE Model} & \multicolumn{3}{c|}{Ferry} & \multicolumn{3}{c|}{Swap} & \multicolumn{3}{c|}{Logistics} & \multicolumn{3}{c|}{Rovers} & \multicolumn{3}{c|}{Floortile} & \multicolumn{3}{c|}{Goldminer} \\
\cline{2-19}
 & APP & PROG & VAL & APP & PROG & VAL & APP & PROG & VAL & APP & PROG & VAL & APP & PROG & VAL & APP & PROG & VAL \\ 
\hline
4o & 100 & 100 & 100 & 100 & 100 & 100 & \textbf{100} & 90 & 70 & \textbf{100} & 90 & \textbf{80} & 90 & 90 & 40 & 90 & 90 & 90 \\
4o-0CoT & 100 & 100 & 90 & 100 & 90 & 100 & 100 & 90 & 80 & 90 & 90 & 80 & \textbf{100} & 80 & \textbf{70} & 60 & 90 & 90 \\
4o-2CoT & 100 & 100 & 90 & 100 & 100 & 100 & 100 & 100 & 60 & 70 & 100 & 60 & 90 & 90 & 70 & 100 & 100 & \textbf{100} \\
4o-SC&70&80&80&80&50&70&50&90&70&60&80&60&50&80&40&60&80&60\\
\textbf{4o-ProRAC}& \textbf{100} & \textbf{100} & \textbf{100} & \textbf{100} & \textbf{100} & \textbf{100} & 90 & \textbf{100} & \textbf{90} & \textbf{100} & \textbf{100} & 70 & 90 & \textbf{90} & 60 & \textbf{100} &\textbf{100}  & 70 \\ 
\hline
4o-mini & 90 & 100 & \textbf{90} & 60 & 100 & 80 & \textbf{100} & 80 & \textbf{80} & \textbf{100} & 90 & 80 & \textbf{100} & \textbf{100} & 60 & 100 & 90 & 80 \\
4o-mini-0CoT & 100 & 100 & 60 & 70 & 100 & 70 & 100 & \textbf{90} & 50 & 80 & \textbf{100} & 80 & 90 & 100 & 40 & 100 & \textbf{100} & \textbf{90} \\
4o-mini-2CoT & 100 & 100 & 80 & \textbf{90} & 100 & 50 & 100 & 80 & 70 & 80 & 90 & \textbf{90} & 100 & 100 & 20 & 100 & 90 & 40 \\
4o-mini-SC&60&80&80&40&60&50&40&90&50&60&90&40&50&80&60&70&90&50\\
\textbf{ 4o-mini-ProRAC} & \textbf{100} & \textbf{100} & 60 & 60 & \textbf{100} & \textbf{90} & 90 & 80 & 70 & 70 & 80 & 70 & 80 & 90 & \textbf{60} & \textbf{100} & 80 & 70 \\ 
\hline
v3 & 100 & 100 & 90 & 100 & 100 & 90 & 100 & 100 & 80 & 80 & 90 & \textbf{100} & 100 & 90 & \textbf{100} & 100 & \textbf{100} & \textbf{90} \\
v3-0CoT & 100 & 100 & 100 & 100 & 100 & \textbf{100} & 100 & 100 & 70 & 90 & 90 & 80 & 100 & 80 & 60 & 90 & 100 & 90 \\
v3-2CoT & 100 & 100 & 100 & 100 & 100 & 100 & 100 & 80 & \textbf{90} & 80 & 90 & 60 & 90 & \textbf{100} & 80 &100  & 100 & 90 \\
v3-SC&60&70&70&40&60&60&50&90&70&70&80&80&60&80&50&70&70&80\\
\textbf{v3-ProRAC}& \textbf{100} & \textbf{100} & \textbf{100} & \textbf{100} & \textbf{100} & 90 & \textbf{100} & \textbf{100} & 70 & \textbf{100} & \textbf{100} & 70 & \textbf{100} & 90 & 60 & \textbf{100} & 80 & 70 \\ 
\hline
\end{tabular}
}
\caption{Performance(\%) of ProRAC and baseline methods on the ACPBench, BOOL questions. APP=Applicability, PROG=Progression, VAL=Validation.}
\label{tab:full-results-ACP-bool}
\end{table*}

For comparison, we adopted Zero-Shot, Zero-Shot-CoT \cite{0shotCOT}, Two-Shot-CoT \cite{FewshotCOT}, and Self-Consistency \cite{SC} prompting as baselines. We selected CoT-based prompting because its step-by-step reasoning paradigm aligns with the core idea of Progression-based reasoning in ProRAC, both involve explicitly modeling intermediate world states to reach the final answer. This conceptual similarity enables a fair and informative comparison. 
Additionally, we included Self-Consistency, which builds on Zero-Shot-CoT by sampling five reasoning paths and selecting the most frequent answer.
In Two-Shot-CoT prompts, we followed a process similar to ProRAC: first identifying the corresponding objects, then checking the executability of actions, subsequently executing the actions to obtain the final state, and finally comparing the query with the final state to determine the answer. All methods were run with temperature set to 0, and one output was generated per input, except for Self-Consistency.

\subsection{Main Results}

Table \ref{tab:full-results-TRAC} presents the results of ProRAC and baseline methods on TRAC. While the LLM performs reasonably well on TRAC using pure prompting methods, its performance improves markedly when enhanced with ProRAC. Considering that the TRAC benchmark involves only the Blocksworld domain and the action sequences are relatively short (specifically, no longer than five steps), it is relatively simple in terms of difficulty. We are more interested in the performance of ProRAC on ActionReasoningBench, where action sequences can be as long as 19 steps.

The performance of ProRAC and baseline methods on ActionReasoningBench is shown in Table \ref{tab:full-result-ARB-1} and Table \ref{tab:full-result-ARB-2}. As observed from the tables, neither the Zero-Shot nor the CoT methods achieved satisfactory results, whereas ProRAC attained significantly better performance across different domains and tasks, even reaching 100\% accuracy in some cases.  In some domains, such as Visitall, ProRAC underperforms compared to the baseline methods. We believe this may be due to how connectivity information is represented in the Visitall domain, where the connections between locations are exhaustively described in the initial state. Such a verbose and information-heavy input poses challenges for LLMs in extracting relevant state information and performing accurate state progression based on it.

Table \ref{tab:full-results-ACP-bool} presents the performance of ProRAC and baseline methods on ACPBench. It can be observed that LLMs perform relatively well on this benchmark, which we primarily attribute to the fact that some of the tasks in ACPBench, such as applicability and progression, are relatively easier than those in the other two benchmarks. However, neither ProRAC nor the baseline methods perform particularly well on the validation task, and their performance in certain domains such as the Logistics and the Floortile declines compared to their performance on the other two tasks. The validation task requires the LLM to distinguish whether an action sequence is valid, applicable, or constitutes a plan. In our experiments, we observed that determining the validity of an action sequence poses a particular challenge for LLMs and they occasionally misclassify valid action sequences as invalid.

\textbf{Experimental results show that our framework is effective.}  ProRAC demonstrates strong performance across different domains and tasks, outperforming prompt-based methods in many scenarios. In certain scenarios, it even achieves 100\% accuracy. The results across all three benchmarks validate the robustness and effectiveness of our approach. One possible reason ProRAC is effective is that it decomposes a question that requires multi-step reasoning in an action-wise manner. Unlike CoT prompting, which instructs the LLM with all intermediate steps in the prompt, ProRAC issues a separate call to the LLM for each action. Each prompt includes the domain description, the current state, and the action to be executed. By reducing the number of input tokens, this design alleviates the processing load on the language model during single-step reasoning, leading to improved performance compared to purely prompt-based methods.


\textbf{Labeling errors in ActionReasoningBench.} It is worth mentioning that during the inspection of our experimental results, we discovered several labeling errors in ActionReasoningBench. After carefully reviewing all responses generated by ProRAC and baseline methods on this benchmark, we believe the benchmark may suffer from systemic annotation issues. We manually corrected all the labeling errors we were able to identify and conducted experiments on the revised benchmark. However, due to limited resources and time, we were unable to exhaustively verify all samples in the benchmark. Further details regarding these labeling issues can be found in supplementary material.


\textbf{The performance of the underlying model can affect overall effectiveness of the framework.} Overall, the DeepSeek-V3 model achieves the best performance. Whether compared with its own prompt-based methods or with other models, DeepSeek-V3 consistently delivers strong results across all three benchmarks. However, the overall performance is less satisfactory when GPT-4o-mini is used as the underlying model. In some cases, it even underperforms compared to prompt-based methods. We consider this may be attributed to the model's smaller parameter size and more limited capabilities.

\subsection{Typical Errors and Their Relation to Problems in Classical AI}
When reviewing incorrect answers made by both the baseline methods and ProRAC, we identified and categorized several common types of errors. Interestingly, these error types closely relate to three important problems in classical AI: the Frame Problem, the Ramification Problem, and the Qualification Problem. This observation highlights that these long-standing challenges remain difficult for large language models to handle effectively. In fact, these issues are not only important for reasoning and planning with LLMs but also significant for fields such as robotics and embodied intelligence.


\textbf{LLMs may incorrectly track or forget object states.} We observe that sometimes after executing an action, the states of objects unrelated to that action change. Other times, after executing an action, the states of objects related to that action do not change. In fact, this issue is related to the Frame Problem, which refers to the challenge for AI systems of efficiently representing and reasoning about aspects of the world that remain unchanged by an action. We observe that even though LLMs have massive numbers of parameters, they sometimes still fail to properly update the states of relevant objects.

\textbf{LLMs may overlook the indirect effects of actions.} We observe that when an action affects the states of multiple objects simultaneously, sometimes LLMs only correctly update the states of some of the objects, while ignoring the indirect effects of the action. For example, in the Blocksworld domain, picking up a block on top of another block and placing it on the table causes the picked-up block to be on the table, and the other block that was previously stacked to become clear. However, the LLM only recognizes that the block placed on the table is clear, without realizing that the other block is also clear. The fundamental cause of such errors is essentially what the Ramification Problem describes, demonstrating the importance of efficiently reasoning about the indirect effects of actions.

\textbf{LLMs may disregard domain descriptions from the prompt.} When executing actions, sometimes LLMs fail to incorporate the preconditions and effects specified in the domain description. There are two situations here: the first is that the LLM fails to recognize that all the preconditions of the action to be executed have already been satisfied. The second is that the LLM adds extra preconditions to the action on its own, causing an action that should be executable to be considered non-executable. The second situation is especially pronounced in GPT-4o-mini. We believe this may be because this model has fewer parameters compared to GPT-4o, resulting in weaker overall comprehension abilities. 
Overall, both of these situations are related to the Qualification Problem: it is challenging for AI systems to enumerate all the preconditions of an action.

\textbf{Errors in initial state extraction.} This error was observed in the first step of ProRAC, where the LLM is asked to extract initial state of all objects. In many scenarios, problems contain a large number of objects, and different objects can affect with each other. This requires the LLM to understand the relationships between different objects as a whole. We found that sometimes LLMs fail to grasp these relationships. For example, in the Depots domain, a crate is placed on a pallet. While the LLM correctly recognizes the crate's position on the pallet, it fails to identify that the pallet is not clear when extracting the state of the pallet itself.

\section{Discussion}

Our evaluation focuses on several existing RAC benchmarks, all synthetic and based on structured ``toy'' domains. This reflects the current state of the field, as no comprehensive and realistic RAC benchmarks exist in the NLP community. We view this work as a necessary step toward addressing more complex scenarios. If LLMs cannot reliably solve basic RAC problems in well-defined settings, their success in open-ended, uncertain real-world contexts is even less likely. While these benchmarks have limitations, we hope our findings inspire not only the development of more realistic benchmarks but also new methods, especially neuro-symbolic approaches. 

\section{Conclusion}
In this work, we introduced ProRAC, a neuro-symbolic method targeting RAC problems. To the best of our knowledge, ProRAC is the first framework that can efficiently solve problems in the three RAC-oriented benchmarks: TRAC, ACPBench, and ActionReasoningBench. Experimental results show that ProRAC is capable of handling RAC problems effectively. Our analysis of experimental results indicates that the three classic and important problems in RAC, the Frame Problem, the Ramification Problem, and the Qualification Problem, still pose significant challenges for large language models. We hope our work can raise awareness of these three problems from the NLP community. An important direction for future work is developing efficient frameworks targeting planning problems incoporating progression and search. Besides, another promising direction is to explore how LLMs can be improved to better handle the Frame, Ramification, and Qualification Problems.

\section{Limitations}
Although ProRAC improves LLMs' performances over RAC problems, it faces some limitations. 
First, its reliance on manually crafted examples for input prompts can be labor-intensive.
Second, while processing a problem, ProRAC calls the LLM via API multiple times, so the API usage cost is also a factor to consider.
Furthermore, in our framework, each subsequent step depends on the result of the previous one, so an error in an earlier step may affect the outcomes of the later steps. 
These limitations may be alleviated in the future as the performance of LLMs improves and their usage costs decrease.

\bibliography{aaai25}

\clearpage
\onecolumn
\appendix
\section{Appendix}
\subsection{Complete experiment results of ACPBench}
\label{subsec:full-results-ACP}

Table \ref{tab:full-results-ACP-mcq} reports results on the multiple-choice questions (MCQs) in ACPBench, comparing ProRAC with baseline methods.

\begin{table}[h]
\centering
\renewcommand{\arraystretch}{1.2}  
\resizebox{1.0\linewidth}{!}{%
\LARGE
\begin{tabular}{|>{\centering\arraybackslash}m{0.29\linewidth}|ccc|ccc|ccc|ccc|ccc|ccc|} 
\hline
\multirow{2}{*}{\centering\LARGE Model} & \multicolumn{3}{c|}{Ferry} & \multicolumn{3}{c|}{Swap} & \multicolumn{3}{c|}{Logistics} & \multicolumn{3}{c|}{Rovers} & \multicolumn{3}{c|}{Floortile} & \multicolumn{3}{c|}{Goldminer} \\
\cline{2-19}
 & APP & PROG & VAL & APP & PROG & VAL & APP & PROG & VAL & APP & PROG & VAL & APP & PROG & VAL & APP & PROG & VAL \\ \hline
4o & 90 & 100 & 40 & 100 & 90 & 100 & 80 & 100 & 40 & 40 & 80 & 30 & 90 & 90 & 40 & 90 & 100 & 90 \\
4o-0CoT & 90 & 100 & 40 & 80 & 100 & 60 & 70 & 100 & 50 & 50 & 80 & 40 & 90 & 80 & 60 & 60 & 90 & 50 \\
4o-2CoT & 90 & 100 & 70 & 90 & 100 & 90 & 80 & 100 & 60 & 50 & 80 & 60 & 90 & 90 & 60 & 90 & 80 & 30 \\
\rowcolor{gray!20}  4o-ProRAC & \textbf{100} & \textbf{100} & \textbf{90} & \textbf{100} & \textbf{100} & 90 & \textbf{100} & \textbf{100} & \textbf{80} & \textbf{90} & \textbf{80} & 30 & \textbf{100} & \textbf{100} & 30 &\textbf{90} & 80 & 30 \\ \hline
4o-mini & 90 & 100 & 30 & 70 & 100 & 70 & 60 & 90 & 20 & 70 & 70 & 20 & 80 & 80 & 40 & 100 & 90 & 80 \\
4o-mini-0CoT & 90 & 100 & 30 & 70 & 100 & 70 & 60 & 90 & 20 & 70 & 60 & 20 & 80 & 90 & 30 & 70 & 90 & 40 \\
4o-mini-2CoT & 90 & 100 & 40 & 80 & 100 & 70 & 50 & 100 & 50 & 50 & 80 & 50 & 80 & 90 & 40 & 90 & 90 & 40 \\
\rowcolor{gray!20}  4o-mini-ProRAC & \textbf{100} & \textbf{100} & 30 & 70 & \textbf{100} & 50 & \textbf{90} & \textbf{100} & 20 & 60 & \textbf{80} & 20 & \textbf{90} & 80 & 30 & 90 & \textbf{100} & 40 \\ \hline
v3 & 100 & 100 & 80 & 90 & 100 & 70 & 100 & 100 & 70 & 70 & 80 & 40 & 80 & 90 & 80 & 100 & 100 & 100 \\
v3-0CoT & 100 & 100 & 80 & 100 & 100 & 80 & 80 & 100 & 80 & 70 & 80 & 30 & 100 & 90 & 60 & 100 & 90 & 20 \\
v3-2CoT & 90 & 100 & 80 & 80 & 100 & 100 & 80 & 100 & 80 & 70 & 80 & 40 & 90 & 90 &60  &100  &100  & 30 \\
\rowcolor{gray!20}  v3-ProRAC & \textbf{100} & \textbf{100} & 70 & 90 & \textbf{100} & 70 & \textbf{100} & \textbf{100} & 40 & \textbf{90} & 70 & 10 & 90 & \textbf{90} & 30 & 90 & \textbf{100} & 40 \\ \hline
\end{tabular}
}
\caption{Performance(\%) of ProRAC and baseline methods on the ACPBench, MCQ questions. APP=Applicability, PROG=Progression, VAL=Validation.}
\label{tab:full-results-ACP-mcq}
\end{table}

\subsection{2Shot-CoT Examples}
In 2Shot-CoT prompts, we followed a process similar to ProRAC: first identifying the corresponding objects, checking the executability of actions, then executing the actions to obtain the final state, and finally comparing the query with the final state to determine the answer. Here we present one example, all prompts can be found here: https://anonymous.4open.science/r/ProRAC-8C46/

\begin{tcolorbox}[colback=gray!10, colframe=gray!80, breakable, title= A 2Shot-CoT Example of Depots domain on Action Executablity task]
[Examples]

(Example 1)

\textbf{Current state:} Crate0 is at depot0, on pallet0, and clear. Crate1 is at depot2, on pallet2, and clear. Crate2 is at depot1, on pallet1, and not clear (crate3 on top). Crate3 is at depot1, on crate2, and clear. Hoist0 is at depot0 and available. Hoist1 is at depot1 and available. Hoist2 is at depot2 and available. Pallet0 is at depot0 and has crate0. Pallet1 is at depot1 and has crate2 (with crate3 on top). Pallet2 is at depot2 and has crate1. Truck0 is at depot2. Truck1 is at depot1.

\textbf{Question:} Given the initial condition, the following action is planned to be performed: at depot2, hoist2 lifts crate1 from pallet2. Is it possible to execute it, True or False?

Answer: 

\textbf{Step 1: we can confirm whether this action is executable or not by checking the initial state.}

A hoist can lift a crate from a pallet only if following preconditions are satisfied: the hoist is available, the crate is clear, and the crate is on the pallet. We can see Hoist2 is at depot2 and available,  Crate1 is at depot2, on pallet2, and clear. So this action is executable.

\textbf{Step 2: we return the final answer based on the checking result.}

Since all preconditions are satisfied, this action is executable.

Final Answer: True

---------------

(Example 2)

\textbf{Current state:} Crate0 is at depot0, on pallet0, and clear. Crate1 is at depot2, on pallet2, and clear. Crate2 is at depot1, on pallet1, but not clear due to crate3 on top. Crate3 is at depot1, on crate2, and clear. Hoist0 is at depot0 and available. Hoist1 is at depot1 and available. Hoist2 is at depot2 and available. Pallet0 is at depot0 and has crate0. Pallet1 is at depot1 and has crate2 with crate3 on top. Pallet2 is at depot2 and has crate1. Truck0 is at depot1. Truck1 is at depot0.

\textbf{Question:} Given the initial condition, the following action is planned to be performed: at depot1, hoist1 lifts crate2 from pallet1. Is it possible to execute it, True or False?

\textbf{Answer: }

\textbf{Step 1: we can confirm whether this action is executable or not by checking the initial state.}

A hoist can lift a crate from a pallet only if the following preconditions are satisfied: the hoist is available, the crate is clear, and the crate is on the pallet.  We can see Hoist1 is at depot1 and available. Crate2 is at depot1, on pallet1, but it is not clear because Crate3 is on top of it. So, this action is not executable

\textbf{Step 2: we return the final answer based on the checking result.}

Since one or more preconditions are violated (specifically, the crate is not clear), the action cannot be performed. 

Final Answer: False.

\end{tcolorbox}

\subsection{General Prompts}
In step one, we extract the initial state, the action sequence, and the query from the problem to be solved. At each step, in addition to the example, we also need to include the instructions that specify what we expect the large model to do as part of the prompt. These instructions are domain-independent and general-purpose.

\begin{tcolorbox}[colback=gray!10, colframe=gray!80, breakable, title=Question Extractor Prompt]
\#This prompt is used instruct the LLM to extract query, which is used in step three.\\

[EXAMPLES]

(EXAMPLE1)

(EXAMPLE2)

----------

[QUESTITION]: ....

Based on examples above, extract the question from the given content. Your answer should meet all the following conditions: First, extract the interrogative sentences and the content following the interrogative sentences, as what exmaples do above. If there is an if conditional clause, also extract the content of the if clause. Second, if the interrogative sentence is followed by the state of the objects, extract that as well. Third, do not use any markdown syntax. Fourth, organize the question you extracred in one paragraph. Remember, you only need to extract the questions, not solve them.
\end{tcolorbox}

\begin{tcolorbox}[colback=gray!10, colframe=gray!80, breakable, title=Executablity Check Prompt]
\#This prompt is used instruct the LLM to check whether this action is executable or not.\\

[EXAMPLES]

(EXAMPLE1)

(EXAMPLE2)

----------

[CURRENT STATE]: ...

[ACTION]: ....

Based on the domain description and examples above, check whether this action is executable at current state. First, think and respond with the format from the example above, including the word choice and paragraph structure. After your analysis, give your final answer in the last paragraph, and the last paragraph should be like "Final Answer: False(True if executable)". Besides, don't use any markdown formatting.
\end{tcolorbox}

\begin{tcolorbox}[colback=gray!10, colframe=gray!80, breakable, title=Action Sequence Extractor Prompt]
\# This prompt is used to instruct the LLM to extract action sequence from the orginal data.\\
This is a plan, you are required to think and understand the entire plan, then extract actions from this plan. Your answer must satisfy all the following requirements. First, Each sentence should contain only one action, and it should be a complete sentence that includes the prepositions and other additional structures from the original plan. Second, don't use any comma in each sentence. Third, Your answer does not need to consider actions that appear in interrogative sentences or actions that appear in if-clauses. Fourth, put all sentences into one paragraph. Remember, you are required to only extract actions, you don't need to solve the question. Besides, don't use any markdown formatting.
\end{tcolorbox}

\subsection{Domain-specific Prompts}
\label{subsec:domain-specific-prompts}
Some prompts, including domain description, executability check, are domain-specific. We present several example prompts used for the Depots domain here. All prompts can be found here: https://anonymous.4open.science/r/ProRAC-8C46/.

\begin{tcolorbox}[colback=gray!10, colframe=gray!80, breakable, title=Domain description of Depots domain]
\# All domain description contains four parts: general description of the domain, predicates for objects, each action's preconditions and each action's effects.\\
Depots domain involves elements such as truck, crate, pallet, distributor, and depot. Hoist and pallet have fixed locations, which can either be at the depot or at the distributor. The location of the crate and truck are not fixed. Crate can be put on a pallet, or put on the top of another crate, or be held by a hoist, or be loaded into a truck. The hoist can only hold one crate at a time. We don't consider crate's stack constraint while it is in the truck.

Different properties are used to describe current states of different objects. For crate we use: at depot/distributor\_X, has crate\_X on top of it/clear, on top of crate\_X/pallet\_X, in truck\_X, held by hoist\_X. For hoist we use: at depot/distributor\_X, available(free)/unavailable, holding crate\_X. For pallet we use: at depot/distributor, has crate\_x on it/clear. For truck we use: at depot/distributor\_X, has crate\_X in it/has no crate in it.

Following actions are executable only if all preconditions are met, if any condition are not satisfied, the action is not executable.
    
A truck can 'drive' from location A to location B. Location can be a depot or distributor. This action is executable only if all following preconditions are satisfied: truck is currently at location A.

A hoist can 'lift' a crate from pallet\_x/crate\_x at the location. This action is executable if all following preconditions are satisfied: the hoist and the crate are both at the same location, the hoist is available, the crate is on the top of pallet\_x/crate\_x, the crate is clear.

A hoist can 'drop' a crate on pallet\_x/crate\_x at the location. This action is executable if all following preconditions are satisfied: the hoist and pallet\_x/crate\_x are both at the same location, pallet\_x/crate\_x is clear, the hoist is holding the crate.

A hoist can 'load' a crate into a truck at the location. This action is executable if all following preconditions are satisfied: the hoist and the truck are at the same location, the hoist is holding the crate.

A hoist can 'unload' a crate from a truck at the location. This action is executable if all following preconditions are satisfied: the hoist and the truck are at the same location, the hoist is available, the crate is on the truck.

Executing an action will change states of related objects.

A truck can 'drive' from location A to location B. This action will result in: the truck is at location B.

A hoist can 'lift' a crate from pallet\_x/crate\_x at the location. This action will result in: the crate is not at the location, the hoist is holding the crate, the hoist is not available, the crate is not clear, pallet\_x/crate\_x is clear, the crate is not on top of pallet\_x/crate\_x.

A hoist can 'drop' a crate on pallet\_x/crate\_x at the location. This action will result in: the hoist is available, the hoist is not holding the crate, the crate is at the location, the crate is clear, the crate is on top of pallet\_x/crate\_x, pallet\_x/crate\_x is not clear.

A hoist can 'load' a crate into a truck at the location. This action will result in: the hoist is available, the hoist is not holding the crate, the crate is in the truck.

A hoist can 'unload' a crate from a truck at the location. This action will result in: the hoist is holding the crate, the hoist is unavailable, the crate is not in the truck.
\end{tcolorbox}

\begin{tcolorbox}[colback=gray!10, colframe=gray!80, breakable, title=Initial state extractor examples of Depots domain]
\textbf{[Examples]}

\textbf{Initial state example 1:} Crate0 can be found located at distributor0, crate0 is clear of any crates, crate0 is on pallet3, crate1 has crate2 on it, crate1 is on top of pallet2, crate3 is at depot2, crate3 is clear, crate3 is on crate2, depot0 is where hoist0 is located, depot1 is where truck1 is located, depot2 is where crate1 is located, depot2 is where crate2 is located, distributor0 is where hoist3 is located, distributor1 is where hoist4 is located, distributor1 is where pallet4 is located, hoist0 is available for work, hoist1 can be found located at depot1, hoist1 is available, hoist2 is available, hoist2 is located at depot2, hoist3 is accessible, hoist4 is available for work, hoist5 is available, hoist5 is located at distributor2, pallet0 is clear, pallet0 is located at depot0, pallet1 is clear, pallet1 is located at depot1, pallet2 can be found located at depot2, pallet3 is located at distributor0, pallet4 is clear of any crates, pallet5 is clear of any crates, pallet5 is located at distributor2, truck0 is at distributor0 and truck2 can be found located at depot0. 

\textbf{Extracted from initial state example 1:}

In order to extract initial states of objects, we first need to confirm how many objects are mentioned in the initial state.

We can find following objects in the intial state description: crate0, crate1, crate2, crate3, hoist0, hoist1, hoist2, hoist3, hoist4, hoist5, pallet0, pallet1, pallet2, pallet3, pallet4, pallet5, truck0, truck1, truck2.

Then, we find all descriptions related to an object, and organize then into the required format. Repeat this process until all objects' states are extracted, and each object's state contains all related properties.

Crate0 can be found located at distributor0, crate0 is clear of any crates, crate0 is on pallet3, ::: Crate0: at distributor0, clear, on top of pallet3. 

crate1 has crate2 on it, crate1 is on top of pallet2, depot2 is where crate1 is located, depot2 is where crate2 is located, crate3 is at depot2, crate3 is clear, crate3 is on crate2,  ::: Crate1: at depot2, on top of pallet2, has crate2 on it. Crate2: at depot2, on top of crate1, has crate3 on it. Crate3: at depot2, on top of crate2, clear. 

depot0 is where hoist0 is located, hoist0 is available for work ::: Hoist0: at depot0, available.

hoist1 can be found located at depot1, hoist1 is available, ::: Hoist1: at depot1, available.

hoist2 is available, hoist2 is located at depot2, ::: Hoist2: at depot2, available.

distributor0 is where hoist3 is located, hoist3 is accessible ::: Hoist3: at distributor0, available.

distributor1 is where hoist4 is located, hoist4 is available for work, ::: Hoist4: at distributor1, available.

hoist5 is available, hoist5 is located at distributor2, ::: Hoist5: at distributor2, available.

truck0 is at distributor0 ::: Truck0: at distributor0, has no crate in it.

depot1 is where truck1 is located, ::: Truck1: at depot1, has no crate in it.

truck2 can be found located at depot0. ::: Truck2: at depot0, has no crate in it.

pallet0 is clear, pallet0 is located at depot0, ::: pallet0: at depot0, clear.

pallet1 is clear, pallet1 is located at depot1, ::: pallet1: at depot1, clear.

pallet2 can be found located at depot2, crate1 is on top of pallet2 ::: pallet2: at depot2, has crate1 on it.

pallet3 is located at distributor0, crate0 is on pallet3, ::: pallet3: at distributor0, has crate0 on it.

pallet4 is clear of any crates, distributor1 is where pallet4 is located, ::: pallet4: distributor1, clear.

pallet5 is clear of any crates, pallet5 is located at distributor2, ::: pallet5: distributor2, clear.

After extracting all objects' state, organize the answer into a new paragraph as the end of answer.

Crate0: at distributor0, clear, on top of pallet3. Crate1: at depot2, on top of pallet2, has crate2 on it. Crate2: at depot2, on top of crate1, has crate3 on it. Crate3: at depot2, on top of crate2, clear. Hoist0: at depot0, available. Hoist1: at depot1, available. Hoist2: at depot2, available. Hoist3: at distributor0, available. Hoist4: at distributor1, available. Hoist5: at distributor2, available. Truck0: at distributor0, has no crate in it. Truck1: at depot1, has no crate in it. Truck2: at depot0, has no crate in it. pallet0: at depot0, clear. pallet1: at depot1, clear. pallet2: at depot2, has crate1 on it. pallet3: at distributor0, has crate0 on it. pallet4: distributor1, clear. pallet5: distributor2, clear.
\end{tcolorbox}

\subsection{Labeling error in ActionReasoningBench}
\label{subsec:labeling-error}
We identified several types of labeling errors in ActionReasoningBench. For the action executability task, some samples contain an action sequences where one of steps is not executable, yet the label is marked as true (or vice versa). For the remaining three tasks, there are cases where, after executing the action sequence, the resulting final state does not satisfy certain propositions in the query, but the label is still marked as true (or vice versa). Here we present a sample from the benchmark, step two of the action sequence is not executable because hoist1 is available and not lifting any crate in the initial state and step one doesn't change hoist1's state.
\begin{tcolorbox}[colback=gray!10, colframe=gray!80, breakable, title=An example with labeling error]
"question\_id": "d9f288db-6871-4608-be0a-0a6408599ee7", 

"domain\_name": "depots", 

"instance\_id": "Instance\_3", 

"question\_category": "action\_executability", 

"question\_name": "iter\_1\_question\_1", 

"fluent\_type": "all\_fluents", 

"answer\_type": "true\_false\_answer", 

"question": "Given the initial condition, the following actions are planned to be performed: at depot0, hoist0 lifts crate2 off pallet0, \textcolor{red}{\textbf{at depot1, hoist1 drops crate2 on pallet1}}, at distributor0, hoist3 unloads crate0 from truck0, at distributor2, hoist5 lifts crate0 off pallet5, at distributor2, hoist5 loads crate1 into truck0, crate1 is lifted from crate0 at distributor2 by hoist5, crate2 is unloaded by hoist1 from truck1 at depot1, crate3 is loaded by hoist2 into truck2 at depot2, from depot0, truck1 is driven to depot1, from depot2, truck2 is driven to distributor3, hoist0 loads crate2 into truck1 at depot0, hoist2 lifts crate3 from pallet2 at depot2, hoist3 drops crate0 on pallet3 at distributor0, hoist5 loads crate0 into truck0 at distributor2, hoist5 unloads crate1 from truck0 at distributor2, hoist6 drops crate3 on pallet6 at distributor3, hoist6 unloads crate3 from truck2 at distributor3, truck0 is driven to distributor0 from distributor2 and truck1 is driven to depot0 from depot1. Is it possible to execute it, True or False?", 

\textcolor{red}{\textbf{"answer": "True"} },"plan\_length": 19, 

"initial\_state\_nl": "Crate0 is at distributor2, crate1 is clear of any crates, crate1 is located at distributor2, crate1 is on crate0, crate2 is clear of any crates, crate3 is clear, crate3 is located at depot2, depot0 is where crate2 is located, depot1 is where hoist1 is located, depot1 is where pallet1 is located, depot1 is where truck1 is located, depot2 is where pallet2 is located, depot2 is where truck2 is located, distributor0 is where pallet3 is located, distributor1 is where pallet4 is located, hoist0 is at depot0, hoist0 is available, \textcolor{blue}{\textbf{hoist1 is available for work}}, hoist2 can be found located at depot2, hoist2 is available, hoist3 is accessible, hoist3 is located at distributor0, hoist4 is accessible, hoist4 is located at distributor1, hoist5 can be found located at distributor2, hoist5 is available, hoist6 is at distributor3, hoist6 is available for work, pallet0 can be found located at depot0, pallet0 has crate2 on it, pallet1 is clear, pallet2 has crate3 on it, pallet3 is clear, pallet4 is clear of any crates, pallet5 has crate0 on it, pallet5 is at distributor2, pallet6 is at distributor3, pallet6 is clear and truck0 can be found located at distributor2.",

\#Following contents in this sample is omitted.
\end{tcolorbox}

\end{document}